\documentclass[runningheads]{llncs}
\usepackage{multirow}
\usepackage{multicol}
\usepackage{amsmath,amssymb,amsfonts,nccmath,bm}
\usepackage{graphicx}
\graphicspath{ {images/} }
\usepackage[skip=3pt]{caption}
\usepackage{subcaption}
\usepackage{hyperref}

\begin{document}

\title{Gaussian Embedding of Large-scale\\ Attributed Graphs}

\author{Bhagya Hettige\inst{1} \and Yuan-Fang Li\inst{1} \and Weiqing Wang\inst{1} \and Wray Buntine\institute{Monash University,
Country, email: somename@university.edu} }

\author{Bhagya Hettige\inst{1} \and
Yuan-Fang Li\inst{1} \and
Weiqing Wang\inst{1} \and Wray Buntine\inst{1}}

\authorrunning{B. Hettige et al.}

\institute{Monash University, Australia
\email{\{bhagya.hettige,yuanfang.li,teresa.wang,wray.buntine\}@monash.edu}}

\maketitle              

\begin{abstract}
Graph embedding methods transform high-dimensional and complex graph contents into low-dimensional representations. 
They are useful for a wide range of graph analysis tasks including link prediction, node classification, recommendation and visualization. 
Most existing approaches represent graph nodes as point vectors in a low-dimensional embedding space, ignoring the uncertainty present in the real-world graphs. 
Furthermore, many real-world graphs are large-scale and rich in content (e.g.\ node attributes). 
In this work, we propose GLACE, a novel, scalable graph embedding method that preserves both graph structure and node attributes effectively and efficiently in an end-to-end manner. 
GLACE effectively models uncertainty through Gaussian embeddings, and supports inductive inference of new nodes based on their attributes. 
In our comprehensive experiments, we evaluate GLACE on real-world graphs, and the results demonstrate that GLACE significantly outperforms state-of-the-art embedding methods on multiple graph analysis tasks.

\keywords{graph embedding \and link prediction \and node classification.}
\end{abstract}

\section{Introduction}

Much real-world data can be expressed as graphs, e.g.\ citation networks \cite{g2g,DBLP:graphsage,DBLP:line}, social-media networks \cite{social1}, language networks \cite{DBLP:PTE,DBLP:line}, and knowledge graphs \cite{DBLP:kg1}.
Graph embedding methods transform graph nodes with highly sparse, high-dimensional content into low-dimensional representations. 
They are effective in capturing complex latent relationships between nodes \cite{DBLP:node2vec,DBLP:graphsage,DBLP:deepwalk,DBLP:line} and have been successfully employed in a wide array of graph analysis tasks such as link prediction, node classification, recommendation and visualization. 
The effective embedding of graph data faces a number of challenges. 

\textbf{Uncertainty modelling:} Most of the previous work~\cite{DBLP:node2vec,DBLP:graphsage,DBLP:deepwalk,DBLP:line} on graph node embedding represents nodes as point vectors in the embedding space, which fails to capture the uncertainty in node representations. Furthermore, graphs constructed from real-world data can be very complex, noisy and imbalanced. Therefore, a mere point-based representation of the nodes may not be able to capture the variability of the graph and so some hidden patterns~\cite{g2g}. 
\textbf{Scalability:} Many real-world graphs are very large, containing millions of nodes and edges. The efficient embedding of such large graphs is thus important but challenging. LINE \cite{DBLP:line} is handling large-scale graphs using an optimized loss function they develop based on local and global network structure, but it does not consider node attributes. 
\textbf{Inductiveness:} Most existing graph embedding approaches are transductive and cannot infer embeddings for nodes unseen at training time. In practice, however, graphs evolve with time, and new nodes and edges can be added into the graph. There are a few recent studies \cite{g2g,DBLP:graphsage} which tried to provide a solution to this limitation. 
However, these methods either do not scale up to large graphs, or require additional information about the graph structure. 

In this paper, we propose GLACE, \textbf{G}aussian representations for \textbf{L}arge-scale \textbf{A}ttributed graph \textbf{C}ontent \textbf{E}mbedding, a novel graph embedding method that addresses all of the above challenges. GLACE learns node embeddings as probability distributions from both node attributes and graph structure information in an end-to-end manner: we use node attributes to initialize the structure-based loss function, and update and transfer the learning back to the encoding function to minimize the loss. 
We use a proximity measure to quantify graph properties to be preserved in the embedding space, i.e.\ first-order proximity to learn from observed relations and second-order proximity to learn from a node's neighbourhoods. 
We learn from node attributes by a non-linear transformation function (encoder), and then define Gaussian embedding functions to model the uncertainty of the embedding by feeding the encoded representation. Therefore, the mean vector of the representation denotes the position of the node in the embedding space, while the covariance matrix gives the uncertainty of the node embedding. 
We deal with new nodes by learning from node attributes, so that a learned model can be used to infer embeddings for new nodes based on their attributes. 
The combination of node attributes and local sampling allows GLACE to be scalable, being able to support graphs of hundred thousand nodes with hundred thousand attributes and half a million edges on modest hardware. 
GLACE derives embeddings from node attributes, which allows it to converge faster during training. 
The main contributions of this work:
\textbf{(1)}~we propose a novel, end-to-end method to embed nodes as probability distributions to model \textbf{uncertainty} of the embedding, 
\textbf{(2)}~our model is \textbf{inductive} as it can infer embedding for unseen nodes using node attributes,
\textbf{(3)}~our model is \textbf{scalable} and \textbf{efficient}, and supports graphs with hundreds of thousands of nodes on modest hardware with a fast convergence rate, while other methods require significantly more memory, more time, or both,
and
\textbf{(4)}~we perform extensive experiments on real-world datasets for link prediction, node classification, induction, and visualization, and GLACE significantly outperforms the baselines.

\section{Related Work}

Below we give a brief overview of recent graph embedding techniques. A more extensive introduction to the area can be found in these recent survey studies \cite{DBLP:survey1,DBLP:survey3,DBLP:survey2}.

Unsupervised graph embedding approaches attempt to preserve graph properties in the embedding space. Random walk-based methods such as DeepWalk~\cite{DBLP:deepwalk} and node2vec~\cite{DBLP:node2vec} generate random walks for each node, and learn embeddings using these node sequences with a technique similar to Skip-Gram~\cite{mikolov2013distributed}. LINE~\cite{DBLP:line} learns from proximity measures considering first- and second-order proximity. SDNE~\cite{DBLP:sdne} proposes a semi-supervised model, in which they learn first-order proximity in the supervised component and second-order proximity in the unsupervised component. Graph2Gauss~\cite{g2g} proposes a personalized ranking scheme such that for a given anchor node, nodes in the immediate neighborhood are closer in the embedding space, while nodes multiple hops away are placed increasingly more distant to the node. Variational graph auto-encoders (VGAE)~\cite{DBLP:gae} is also an unsupervised learning method for undirected graphs.

Learning \emph{uncertainty} of embeddings has been shown to produce meaningful representations \cite{DBLP:word2gauss,g2g}. Word2gauss~\cite{DBLP:word2gauss} proposes a Gaussian embedding space to model word embeddings. Graph2Gauss~\cite{g2g} captures uncertainty of graph nodes similarly. Both methods show that capturing embedding uncertainty learns more meaningful representations in their evaluation tasks. Another recent study~\cite{DBLP:vne} proposes to learn node embeddings as Gaussian distributions using the Wasserstein metric rather than KL divergence, as the former preserves edge transitivity.

Graphs can vary greatly in \emph{size} (i.e.\ number of nodes and edges). Some methods are designed to be scalable while others do not scale well due to high space and/or time complexities. LINE \cite{DBLP:line} is a method designed to handle large-scale graphs efficiently using negative sampling and edge sampling optimization strategies. Graph2Gauss \cite{g2g}, on the other hand, exhibits poor scalability as it needs to compute hops for each node up to a predefined number. This process is not only time consuming, but also consumes significant memory. 

\section{GLACE Methodology}

\subsection{Notations and Problem Definition}

\textbf{Homogeneous Graph:} Let $G = (\mathcal{V}, \emph{E}, \mathbf{X})$ be an attributed homogeneous graph, where $\mathcal{V}$ is the set of nodes, $\emph{E}$ is the set of edges between nodes in $\mathcal{V}$, where each ordered pair of nodes $(i, j) \in \emph{E}$ is associated with a weight $w_{ij} > 0$ for edge from $i$ to $j$, and $\mathbf{X}\in\mathbb{R}^{|\mathcal{V}| \times D}$ is the attribute matrix of the nodes which represents an attribute matrix of $\mathcal{V}$, where $\textbf{x}_i\in\mathbf{X}$ is a $D$-dimensional attribute vector of node $i$.

\noindent
\textbf{GLACE Embedding:} GLACE aims to represent each node $i \in \mathcal{V}$ as a low-dimensional Gaussian distribution embedding, $\mathbf{z}_i = \mathcal{N}(\mu_{i}, \Sigma_{i})$, where $\mu_{i} \in \mathbb{R}^L$, $\Sigma_{i} \in \mathbb{R}^{L \times L}$ where $L$ is the embedding dimension with $L \ll |\mathcal{V}|, D_k$, in embedding space $\mathbb{R}^L$, such that nodes close to each other in the original graph are also close in the embedding space.
We learn $\Sigma_i$ as a diagonal covariance vector, $\Sigma_i \in \mathbb{R}^L$, instead of a covariance matrix to reduce the number of parameters to learn.

\subsection{Overall Architecture}

GLACE is an end-to-end framework for learning node embeddings using both node attributes and graph structure in an efficient manner. 
Node attributes are first fed through a non-linear transformation function and then through two non-linear transformation functions to obtain a mean vector and diagonal covariance vector which represent a Gaussian embedding. 
GLACE is flexible in handling different node attribute formats, such as text and images, since we can define the encoder architecture accordingly.
Our unsupervised loss function is defined based on graph structure. 
We learn local and global graph structure using our proximity measure, since we can optimize the function using negative sampling \cite{mikolov2013distributed} to achieve scalability.
Local structure is learnt with first-order proximity, i.e. based on edge weight between nodes \cite{DBLP:survey1}, and global structure is learnt with second-order proximity, i.e. based on similarity between neighborhoods of a pair of nodes \cite{DBLP:survey1}. 
GLACE learns in an end-to-end manner: 
\textbf{forward learning:} we use encoded node attributes as input to the optimization function of Graph Structure Encoding, and 
\textbf{back-propagation:} we minimize the optimization function of Graph Structure Encoding by updating the node embeddings, and then propagating the update back to the Node Attribute Encoding part.

\subsection{Node Attribute Encoding}
\label{sec:node_attr}

We learn node attributes using two levels of transformations, encoding and Gaussian embedding. 
At the first level, we use a multi-layer perceptron (MLP) to encode the node attribute information and generate an intermediate vector from node attribute information. 
We use a feed-forward encoder, $f: \mathcal{V} \to \mathbb{R}^m$ which takes an attribute vector $\mathbf{x}_{i} \in \mathbf{X}$ as input for node $i$, and outputs a $m$-dimensional intermediate vector.
\useshortskip
\begin{equation}\small
    \label{eq:enc}
    \mathbf{u}_i = f(\mathbf{x}_{i}) = \mathbf{W}\mathbf{x}_{i} + \mathbf{b}
\end{equation}
The attribute encoder of the model is expressed using weight matrix $\mathbf{W} \in \mathbb{R}^{D \times m}$ and bias vector $\mathbf{b} \in \mathbb{R}^{m}$ where $m$ is the dimension of the hidden representation. 
Note here that, we can easily alter the encoder architecture such that it aligns and captures different types of inputs (e.g.\ images, text).
But for efficiency purposes we have only considered an MLP architecture. 
This intermediate vector $\mathbf{u}_i$ is then used as input to two encoders $f_\mu$ and $f_\Sigma$ to learn $\bm{\mu}$ and $\mathbf{\Sigma}$ in the Gaussian distributions. 
The final latent representation of node $i$ of type $k$ is $\mathbf{z}_i = \mathcal{N}(\bm{\mu}_{i}, \mathbf{\Sigma}_{i})$, where $\bm{\mu}_{i} = f_\mu(f(\mathbf{x}_{i}))$ and $\mathbf{\Sigma}_{i} = f_\Sigma(f(\mathbf{x}_{i}))$.
{\small
  \begin{align}
    \label{eq:mu}
    \bm{\mu}_i = f_\mu(\mathbf{u}_i) &= \mathbf{W}_\mu\mathbf{u}_i + \mathbf{b}_\mu\\
    \label{eq:sigma}
    \mathbf{\Sigma}_i = f_\Sigma({\mathbf{u}_i}) &= ELU(\mathbf{W}_\Sigma\mathbf{u}_i + \mathbf{b}_\Sigma) + 1
  \end{align}
}%

The two functions defined in Eq.~\ref{eq:mu} with $\mathbf{W}_\mu \in \mathbb{R}^{m \times L}$ and $\mathbf{b}_\mu \in \mathbb{R}^{L}$, and in Eq.~\ref{eq:sigma} with $\mathbf{W}_\Sigma \in \mathbb{R}^{m \times L}$ and $\mathbf{b}_\Sigma \in \mathbb{R}^{L}$ denote the Mean Encoder and Covariance Encoder respectively. Note that, as the difference between different node types have been caught by $\mathbf{u}_i$ generated by $f_k$, all the node types share the same Mean Encoder and Covariance Encoder in GLACE to achieve good scalability.  
Here for the uncertainty representation, we constrain the covariance matrix to be diagonal to reduce the number of parameters to learn. The exponential linear unit (ELU) 
is used as the activation function in the Covariance Encoder. An ELU can have negative values as well, and it drives the mean of the activation outputs be closer to zero which makes learning and convergence much faster. We add 1 to obtain positive covariance. 

Note that, even inside the Node Attribute Encoding component, GLACE also learns the parameters in an end-to-end manner. Through the shared parameter $\mathbf{u}_i$, GLACE forwards the updating inside Encoder $f_k$ to Gaussian Encoders $f_\mu$ and $f_\Sigma$, and propagates the updating inside Gaussian Encoders back to $f_k$ automatically during the optimization process. 

\subsection{Graph Structure Encoding}
\label{sec:graph_struct}

GLACE aims at capturing both local (first-order) and global (higher-order) proximity information in graphs. But considering the scalability to large-scale graphs, for the global information, GLACE only encodes second-order proximity. For each node $i$, the learned Gaussian distributions, $\mathbf{z}_i$, in Section \ref{sec:node_attr} are used as the input to the Graph Structure Encoding component in this section.  

\textbf{Dissimilarity measure:} Let $d(\mathbf{z}_i, \mathbf{z}_j)$ be the dissimilarity measure between latent representations of two nodes $i, j \in \mathcal{V}$. 
Since $\mathbf{z}_i$ and $\mathbf{z}_j$ are Gaussian distribution embedding, we should select a dissimilarity measure to be a function to \textit{measure the dissimilarity between two probability distributions}. 
Therefore, the dissimilarity measure between two latent representations is calculated using asymmetric KL divergence, $d(\mathbf{z}_i, \mathbf{z}_j) = D_{KL}(\mathbf{z}_j || \mathbf{z}_i)$.
Alternatively, we could also use a Wasserstein metric instead of KL divergence as in \cite{DBLP:vne}.
Since KL divergence is asymmetric, for undirected graphs we extend the distance to a symmetric dissimilarity measure as:
\useshortskip
\begin{equation}\small
    d(\mathbf{z}_{i}, \mathbf{z}_{j}) = \frac{1}{2}(D_{KL}(\mathbf{z}_i || \mathbf{z}_j) + D_{KL}(\mathbf{z}_j || \mathbf{z}_i))
    \label{eq:kl}
\end{equation}

\textbf{First-order proximity:} We learn first-order proximity of nodes, by modelling local pairwise proximity between two connected nodes in the graph. The empirical probability for first-order proximity measure observed in the original graph between nodes $i$ and $j$ is defined as the ratio of the weight of the edge $(i, j)$ to the total of the weights of all the edges in the graph. For each \textit{undirected} edge $(i, j)$ we define the joint probability as a sigmoid function between node $i$ and $j$. These two functions can be defined as respectively:
\useshortskip
\begin{equation}\small
    \label{eq:p1}
    {\hat{p}}_1(i, j) = \frac{w_{ij}}{\Sigma_{(\hat{i}, \hat{j}) \in \emph{E}} w_{\hat{i}\hat{j}}} \text{   and   } p_1(i, j) = \frac{1}{1 + \exp{(d(\mathbf{z}_i, \mathbf{z}_j))}}
\end{equation}
We preserve the first-order proximity by minimizing the distance between the two distributions, $O_1 = D_{KL}(\hat{p}_1 || p_1)$, for all edges. Motivated by this function, we use the following objective function as in LINE~\cite{DBLP:line} for first-order proximity:
\useshortskip
\begin{equation}\small
    \label{eq:o1}
    O_1 = - \sum_{(i,j) \in \emph{E}} w_{ij} \log p_1(i,j)
\end{equation}
 
\textbf{Second-order proximity:}
Nodes which have more similar neighbourhoods should be closer in embedding space with respect to the nodes with less similar neighbourhoods. The empirical probability of second-order proximity observed for edge $(i, j)$ can be defined as the ratio of the weight of edge $(i, j)$ to the total weight of edges from node $i$ to its immediate neighborhood, $N(i)$. Similarly to LINE, each node is represented with two complementary embeddings, the first embedding $\mathbf{z}_i$, is as defined previously, and the second is the context embedding, $h^\prime_{i}$, defined in Eq.~\ref{eq:mu_cnx} and Eq.~\ref{eq:sigma_cnx}. For each \textit{directed} edge $(i, j)$ (if the edge is undirected, it can be treated as two edges with equal weights and opposite directions) we define the the probability of \textit{context} $j$ generated by node $i$ as a softmax function. The two probability distributions are defined as follows:
\useshortskip
\begin{equation}\small
    \label{eq:p2}
     \hat{p}_2(j | i) = \frac{w_{ij}}{\Sigma_{\hat{i} \in N(i)} w_{i\hat{i}}} \text{   and   } p_2(j | i) = \frac{\exp{(-d(\mathbf{z}_i, \mathbf{z}^\prime_{j}))}}{\Sigma_{\hat{i} \in V} \exp{(-d(\mathbf{z}_i, \mathbf{z}^\prime_{\hat{i}}))}}
\end{equation}
We preserve the second-order proximity by minimizing the distance between the two distributions, $O_2 = \sum_{i \in \mathcal{V}} \lambda_{i} D_{KL}(\hat{p}_2(.|i) || p_2(.|i))$, where $\lambda_{i}$ is the prestige of node $i$. Motivated by this~\cite{DBLP:line} we preserve the second-order proximity:
\useshortskip
\begin{equation}\small
    \label{eq:o2}
    O_2 = - \sum_{(i,j) \in E} w_{ij} \log p_2(i,j)
\end{equation}

When we define the second-order proximity measure, the neighbourhood nodes are considered as contexts for the anchor node. Therefore, we should define another set of node attribute encoding functions to model the context representations used for neighbourhood nodes, similarly to the Equations: \ref{eq:enc}, \ref{eq:mu} and \ref{eq:sigma}. The encoder for context nodes is $f^\prime: \mathcal{V} \to \mathbb{R}^m$. The latent representation of context node $i$ is $\mathbf{z}^\prime_i = \mathcal{N}(\bm{\mu}^\prime_{i}, \bm{\Sigma}^\prime_{i})$, where $\bm{\mu}^\prime_{i} = f_\mu^\prime(f^\prime(\mathbf{x}_{i}))$ and $\Sigma^\prime_{i} = f_\Sigma^\prime(f^\prime(\mathbf{x}_{i}))$.
{\small
  \begin{align}
    \label{eq:enc_cnx}
    \mathbf{u}^\prime_i = f^\prime(\mathbf{x}_{i}) &= \mathbf{W}^\prime\mathbf{x}_{i} + \mathbf{b}^\prime\\
    \label{eq:mu_cnx}
    \bm{\mu}^\prime_i = f_\mu^\prime(\mathbf{u}_i) &= \mathbf{W}^\prime_\mu\mathbf{u}_i + \mathbf{b}^\prime_\mu\\
    \label{eq:sigma_cnx}
    \mathbf{\Sigma}^\prime_i = f_\Sigma^\prime(\mathbf{u}_i) &= ELU(\mathbf{W}^\prime_\Sigma\mathbf{u}_i + \mathbf{b}^\prime_\Sigma) + 1
  \end{align}
}%

\subsection{Model Optimization}

The objective function in Eq.~\ref{eq:o2} is a bottleneck as it requires evaluation on the entire set of nodes for the optimization of one single edge as shown in Eq.~\ref{eq:p2}. Based on the negative sampling approach~\cite{mikolov2013distributed,DBLP:line}, we sample several negative edges (i.e., defined as $N$) for each edge in the training set to optimize the objective function. With negative sampling our objective function $O_2$ in Eq.~\ref{eq:o2} becomes:
\useshortskip
\begin{equation}\small
    \label{eq:neg_sample}
    \sum_{(i,j) \in \emph{E}} \big( \log\;{\sigma(-d(\mathbf{z}_i, \mathbf{z}_j^\prime))} + \sum_{n = 1}^{N} \mathbb{E}_{v_n \thicksim P_{n}(v)} \log \;{\sigma(d(\mathbf{z}_i, \mathbf{z}^\prime_{v_n}))} \big)
\end{equation}
Similarly, we can efficiently compute $O_1$ in Eg.~\ref{eq:o1} with negative sampling:
\useshortskip
\begin{equation}\small
    \label{eq:neg_sample1}
    \sum_{(i,j) \in \emph{E}} \big( \log\;{\sigma(-d(\mathbf{z}_i, \mathbf{z}_j))} + \sum_{n = 1}^{N} \mathbb{E}_{v_n \thicksim P_{n}(v)} \log \;{\sigma(d(\mathbf{z}_i, \mathbf{z}_{v_n}))} \big)
\end{equation}
where we draw negative edges from the noise distribution $P_{n}(v)$ with negative node probability distribution, $P_{n}(v) \propto out\_degree(v)^{3/4}$ for $v \in \mathcal{V}$. Similarly, we can optimize objective function $O_1$ in Eq.~\ref{eq:o1}, replacing $\mathbf{z}_j^\prime$ and $\mathbf{z}^\prime_{v_n}$ in Eq.~\ref{eq:neg_sample} with $\mathbf{z}_j$ and $\mathbf{z}_{v_n}$ respectively.
We further optimize our training process by implementing early stopping for training algorithm using a validation set and assessing the performance at each iteration. 

\subsection{Complexity Analysis}

Training of GLACE takes $O(T \times b \times (dN + (N+2) \times (Dm + mL + L))) = O(T \times b \times N \times (d + Dm + mL + L))$, where $T$ is the maximum number of iterations, $b$ is the batch size, $d$ is the maximum node degree, $N$ is the number of negative samples, $D$ is the attribute vector dimension, $m$ is the intermediate vector dimension (hidden layer of Node Attribute Encoder), and $L$ is the embedding dimension. For each edge in the batch, fetching $N$ negative samples takes $O(dN)$ time. For each of the $(N+2)$ nodes, i.e., $i, j$ and $\{v_n\}_{v_n \in Neg(i)}$, we compute and update parameters in the Node Attribute Encoder with two levels of transformations (i.e., $f_{enc}, f_\mu$ and $f_\Sigma$) in $O(Dm) + O(2mL) + O(2L) = O(Dm + mL + L)$ time. 
Since GLACE initializes node embeddings using encoded node attribute information, it can achieve faster convergence in optimization (in practice we can see that GLACE starts to reach optimization point at $T=100$. We will discuss further our method's scalability over LINE in the experiments section). 

\section{Experiments}
\label{sec:experiments}

We evaluate our method with state-of-the-art baselines on: link prediction, node classification, inductive learning and visualization.
In addition, we demonstrate the scalability and inductiveness of our model. 
Source code for GLACE is publicly available at \url{https://github.com/bhagya-hettige/GLACE}.

\subsection{Datasets}

We use six publicly available real-world attributed graphs (Table \ref{tab:homo_data_stat}).
These are citation networks in which nodes denote papers and edges represent citation relations. 
For each paper, we have TF-IDF vectors of the paper's abstract as attributes.
Cora-ML is a subset extracted from the Cora citation network. 
The larger ACM network
is constructed using Aminer data~\cite{aminer_data}. 

\begin{table}[t]
\caption{\small Statistics of the real-world graphs.}
\label{tab:homo_data_stat}
\vspace{-2mm}
\begin{center}
    \small
    \begin{tabular}{|l|c|c|c|c|}
        \hline
        \textbf{Dataset} & \textbf{$|\mathcal{V}_1|$} & $|\emph{E}|$ & \textbf{$D_1$} & \textbf{\#Labels}\\
        \hline
        Cora-ML & 2,995 & 8,416 & 2,879 & 7 \\
        Cora & 19,793 & 65,311 & 8,710 & 70 \\
        Citeseer & 4,230 & 5,358 & 2,701 & 6 \\
        DBLP & 17,716 & 105,734 & 1,639 & 4 \\
        Pubmed & 18,230 & 79,612 & 500 & 3 \\
        ACM & 115,772 & 539,910 & 124,856 & - \\
        \hline
    \end{tabular}
\end{center}
\vspace{-8mm}
\end{table}

\subsection{Compared Algorithms and Setup}

All the experiments were performed on a MacBook Pro laptop with 16GB memory and a 2.6 GHz Intel Core i7 processor. For each of the following models, we give maximum of 5 hours as a threshold for training.

\textbf{Attributes}: for evaluation tasks, we use raw node attributes as input features instead of node embeddings.
\textbf{node2vec} \cite{DBLP:node2vec}: is a random walk based node embedding method that maximizes the likelihood of preserving nodes' neighbourhood using biased random walks starting from each node. Therefore, node2vec considers only second-order proximity.
\textbf{LINE} \cite{DBLP:line}: is for large-scale non-attributed graphs and uses first-order and second-order proximity information.
\textbf{GraphSAGE} \cite{DBLP:graphsage}: is an inductive learning approach for attributed graphs which learns an embedding function by sampling and aggregating features of local neighbourhoods of nodes. We use the unsupervised version of GraphSAGE with the pooling aggregator (which performed best for citation networks according to \cite{DBLP:graphsage}). Since we use node class labels in the node classification task, supervised version of GraphSAGE is not considered in evaluation.
\textbf{Graph2Gauss (G2G)} \cite{g2g}: produces Gaussian node embeddings using node attributes and graph structure, which introduces a personalized ranking of nodes based on neighbouring hops. G2G is applicable to homogeneous graphs with plain/attributed nodes and (un)directed and unweighted edges.

We also include a non Gaussian representation model to assess the effectiveness of uncertainty modelling.
\textbf{LACE} (without Gaussians): We use a version of our method in which we represent nodes as vectors in an embedding space using node attributes and graph structure. 
\textbf{GLACE} (with Gaussians): This is the complete version of our method which produces Gaussian distribution representations for graph nodes using node attributes and graph structure.

For LINE and GLACE, we consider first-order ($1^{st}$), second-order ($2^{nd}$) and a concatenated representation of first- and second-order proximities ($1^{st} + 2^{nd}$).
Accordingly, the concatenated representation would have both local and global information.
For all the models, we use 128 as the dimension of the embedding.
Since GLACE learns two vectors for mean and variance respectively, we set $L=64$ to conduct a fair comparison with other methods, so the number of dimensions learned for each node still remains the same.

\subsection{Link Prediction}

For all the methods we extract a test set containing 20\% randomly selected edges from the graph and an equal number of non-edges which are not present in the graph. 
For all datasets we use the same splits for all the methods. 
The remaining 80\% of the edges are used for training the embedding models.
In probability distribution based embedding methods (G2G and GLACE) we use negative KL divergence to rank the Gaussian embeddings. 
For other embedding methods (attributes, node2vec, LINE and LACE), we use dot product similarity of node embedding to ranking.
We consider both $1^{st}$-order and $2^{nd}$-order proximity.  We also consider joint embedding performance by concatenating the resulting embedding from the two proximity. 
For LINE, we record the concatenated embedding of the two proximities, which is identified as the best-performing setting for LINE~\cite{DBLP:line}. 
AUC and AP scores of link prediction task are shown in Table~\ref{tab:link_prediction_homo}.

A number of important observations can be made from the tables. \textbf{(1)} GLACE clearly outperforms the state-of-art embedding methods by a significant margin in both homogeneous and bipartite graphs. The introduction of uncertainty modelling in GLACE improves performance considerably when compared to models without Gaussian embedding, i.e.\ node2vec, LINE and LACE. \textbf{(2)} In homogeneous graphs, GLACE$_{(1^{st}+2^{nd})}$, which learns from both the explicit edges in the graph and neighbourhood similarities, is the best performing model. \textbf{(3)} G2G shows a very competitive performance to GLACE in smaller graphs (Cora, DBLP and Pubmed) due to its hop-based node ranking scheme, but it does not scale up for large-scale graphs, ACM and Stackoverflow. 
\textbf{(4)} GLACE's better scalability is also shown, as it is the only attributed graph embedding model that completes the largest dataset, ACM.

\begin{table}[t]
\caption{\small Link prediction performance. Experiments not completed within threshold settings are marked with "-".}
\label{tab:link_prediction_homo}
\vspace{-2mm}
\begin{center}
    \begin{tabular}{|l|cc|cc|cc|cc|cc|}
        \hline
        \textbf{Algorithm} &
        \multicolumn{2}{c|}{\textbf{Cora}} & \multicolumn{2}{c|}{\textbf{Citeseer}} & \multicolumn{2}{c|}{\textbf{DBLP}} & \multicolumn{2}{c|}{\textbf{Pubmed}} & \multicolumn{2}{c|}{\textbf{ACM}}\\
        & \textbf{AUC} & \textbf{AP} & \textbf{AUC} & \textbf{AP} & \textbf{AUC} & \textbf{AP} & \textbf{AUC} & \textbf{AP} & \textbf{AUC} & \textbf{AP}\\
        \hline
        Attributes & 
        82.98 & 77.71 & 81.53 & 75.60 & 75.89 & 69.56 & 82.98 & 77.71 & - & - \\
        node2vec & 
        87.86 & 87.19 & 79.91 & 82.08 & 87.03 & 84.36 & 88.74 & 86.58 & 91.18 & 91.49 \\
        LINE & 
        75.23 & 77.96 & 71.20 & 72.11 & 80.01 & 83.09 & 79.97 & 82.86 & 75.32 & 76.81 \\
        GraphSAGE &
        85.30 & 84.72 & 83.33 & 85.38 & 89.63 & 90.12 & 89.43 & 90.90 & - & - \\
        G2G & 
        97.87 & 98.03 & 96.28 & 96.54 & 96.35 & 96.79 & 95.75 & 95.65 & - & - \\
        \hline
        LACE$_{(1^{st})}$ & 
        96.59 & 96.66 & 94.21 & 94.95 & 91.91 & 92.68 & 83.89 & 84.26 & 95.14 & 95.07 \\
        LACE$_{(2^{nd})}$ & 
        96.83 & 96.67 & 94.29 & 94.61 & 93.30 & 93.37 & 93.72 & 92.80 & 94.37 & 93.91 \\
        LACE$_{(1^{st}+2^{nd})}$ & 
        97.51 & 97.40 & 95.35 & 95.76 & 93.82 & 94.14 & 89.53 & 89.85 & 96.01 & 95.79 \\
        GLACE$_{(1^{st})}$ & 
        \underline{98.54} & \underline{98.46} & 96.41 & 96.40 & \underline{98.48} & \underline{98.33} & \underline{97.69} & \underline{97.42} & \underline{98.00} & \underline{97.94} \\
        GLACE$_{(2^{nd})}$ & 
        98.43 & 98.31 & \underline{97.22} & \underline{97.20} & 98.16 & 97.95 & 97.02 & 96.56 & 97.94 & 97.79 \\
        GLACE$_{(1^{st}+2^{nd})}$ & 
        \textbf{98.60} & \textbf{98.52} & \textbf{98.43} & \textbf{98.37} & \textbf{98.55} & \textbf{98.40} & \textbf{97.82} & \textbf{97.49} & \textbf{98.34} & \textbf{98.24} \\
        \hline
    \end{tabular}
\end{center}
\vspace{-8mm}
\end{table}

\subsection{Multi-class Node Classification}

\begin{figure*}[t]
    \begin{minipage}{\textwidth}
        \centering
        \includegraphics[width=\linewidth]{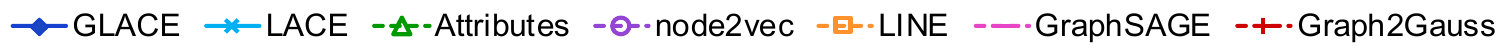}
    \end{minipage}
    
    \begin{minipage}{\textwidth}
        \centering
        \begin{subfigure}{0.32\textwidth}
            \includegraphics[width=\linewidth]{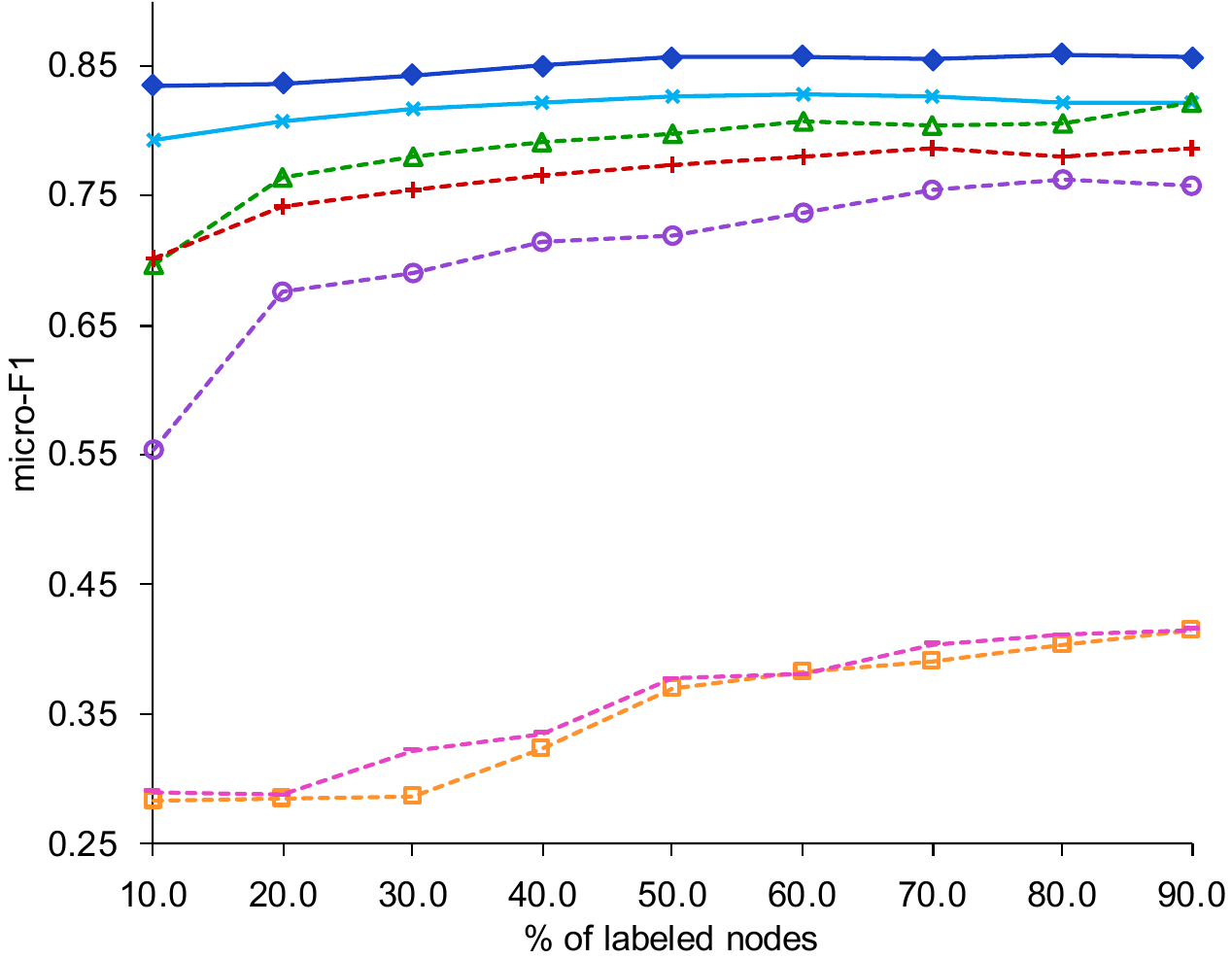}
            \caption{Cora-ML}
            \label{fig:cora_ml}
        \end{subfigure}
        \begin{subfigure}{0.32\textwidth}
            \includegraphics[width=\linewidth]{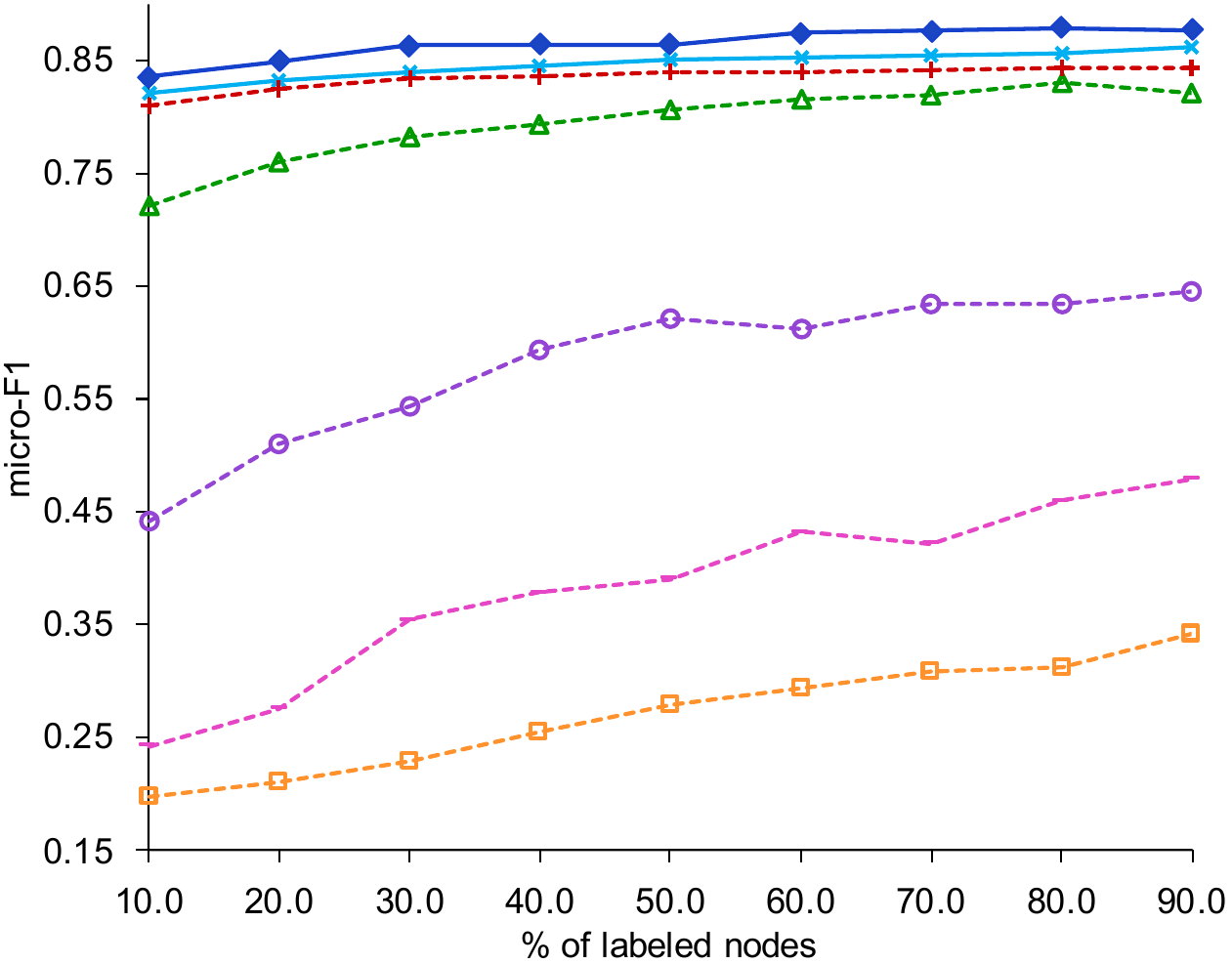}
            \caption{Citeseer}
            \label{fig:citeseer}
        \end{subfigure}
        \begin{subfigure}{0.32\textwidth}
            \includegraphics[width=\linewidth]{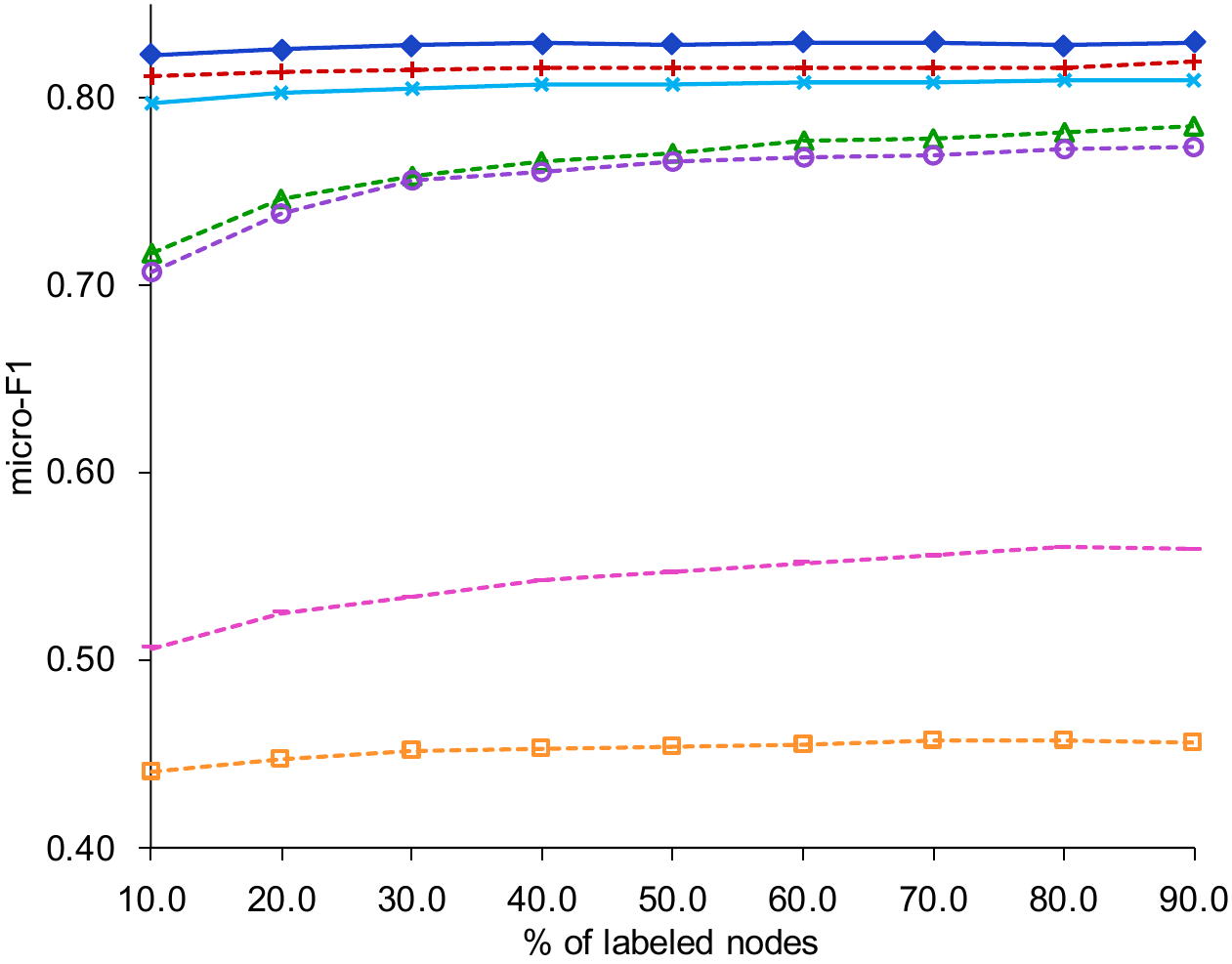}
            \caption{DBLP}
            \label{fig:dblp}
        \end{subfigure}
    \end{minipage}
    \caption{\small Node classification performance. Improvements of GLACE are statistically significant for $p<0.01$ estimated by a paired t-test.}
    \label{fig:node_class}
    \vspace{-5mm}
\end{figure*}

The node embeddings are obtained using the complete node set from the evaluated models. Similarly to \cite{DBLP:line,g2g}, we randomly sample different percentages of labeled nodes from the graph for training a logistic regression classifier to predict class label, and use the rest of the nodes for evaluation. The percentages of nodes used for training the classifier for node classification task are $10\%,20\%,\dots,90\%$. The evaluation metric we report is F1-score, and the results are averaged over 10 trials. We performed this experiment on all the evaluated graphs, and we report the results for Cora-ML, Citeseer, and DBLP datasets in Figure~\ref{fig:node_class}.

Based on the results, it can be seen that GLACE again consistently outperforms the baseline methods. This is clearly due to uncertainty modelling of the representations. As can be seen in the figures, there is a clear separation of node classification performance between the methods that consider node attributes and the methods that do not. An exception to this observation is GraphSAGE, which considers attributes but has a considerably poorer performance than GLACE, G2G and LR. 
This can be due to its aggregation process which magnifies any error.
LACE (without Gaussians) is able to outperform some of the baseline methods, and this is due to the incorporation of node attributes.

\subsection{Inductive Learning}

We have evaluated the inductive property by training the models with 10\% and 50\% nodes hidden from the original graph. Then we evaluate how well the models can infer embeddings for unseen nodes on the link prediction task, comparing our model against G2G~\cite{g2g}, which also takes attributes into account. Although GraphSAGE~\cite{DBLP:graphsage} is also an inductive node embedding method, it is not applicable in this scenario as it requires unseen nodes to be connected to existing nodes. 
We perform this task on Cora-ML, Citeseer, Pubmed and ACM graphs. 
Table \ref{tab:link_prediction_ind} summarizes the results. 

As can be seen from the table, GLACE outperforms G2G across all the datasets over the two hidden percentage values. It can also be observed that, GLACE suffers considerably less performance degradation than G2G when more nodes are hidden (i.e.\ from 10\% to 50\%). Since G2G requires constructing hops and keeping them in memory, we could not run experiments for G2G (with maximum number of hops to consider $>1$) on the ACM dataset, which also demonstrates the scalability advantage of GLACE.

\begin{table}[t]
\caption{\small Link prediction performance with inductive learning.}
\label{tab:link_prediction_ind}
\vspace{-2mm}
\begin{center}
    \begin{tabular}{|l|cc|cc|cc|cc|}
        \hline
        \textbf{Algorithm} &\multicolumn{2}{c|}{\textbf{Cora-ML}} &  \multicolumn{2}{c|}{\textbf{Citeseer}} & \multicolumn{2}{c|}{\textbf{Pubmed}} & \multicolumn{2}{c|}{\textbf{ACM}} \\
        \textbf{[hidden \%]} & \textbf{AUC} & \textbf{AP} & \textbf{AUC} & \textbf{AP} & \textbf{AUC} & \textbf{AP} & \textbf{AUC} & \textbf{AP} \\
        \hline
        G2G [10\%] & 88.83 & 79.34 & 87.96 & 80.39 & 88.96 & 77.08 & - & - \\
        GLACE [10\%]  & 93.07 & 86.72 & 90.76 & 85.03 & 93.00 & 84.19 & 95.05 & 89.09 \\
        \hline
        G2G [50\%] & 57.26 & 34.70 & 61.71 & 43.87 & 51.22 & 27.39 & - & - \\
        GLACE [50\%]  & 87.62 & 74.64 & 83.69 & 70.74 & 92.18 & 79.99 & 93.96 & 85.33 \\
        \hline
    \end{tabular}
\end{center}
\vspace{-8mm}
\end{table}

\subsection{Scalability}

\begin{figure}[t]
    \centering
    \includegraphics[width=0.75\linewidth]{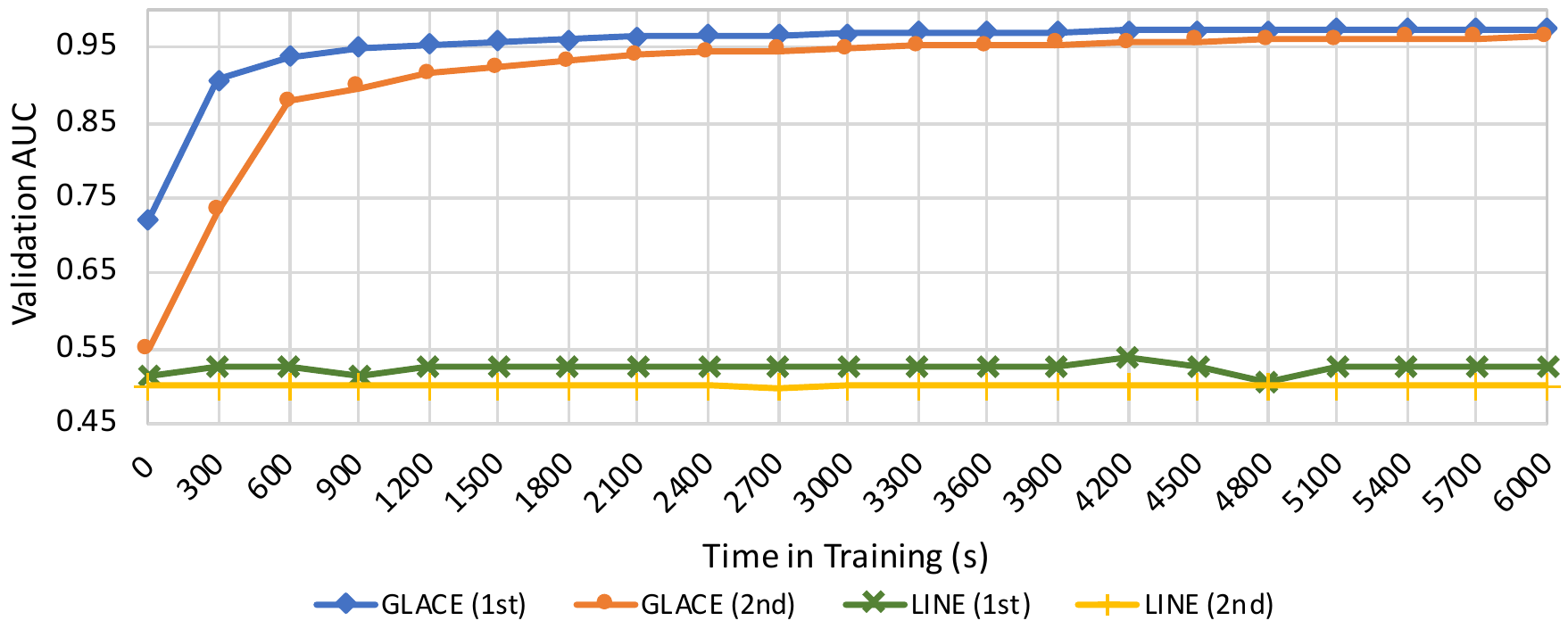}
    \caption{\small GLACE's faster convergence. Link prediction performance in ACM training.}
    \label{fig:scale}
    \vspace{-4mm}
\end{figure}

LINE is a scalable embedding method for plain graphs. In this study we introduced GLACE as an improved scalable embedding method for attributed graphs. In this section we evaluate the efficiency of our method against the large-scale embedding method, LINE, and see how the introduction of attributes and uncertainty modelling assist GLACE in converging faster for optimization. We report the validation AUC for link prediction task in ACM dataset against time. The trend is similar in other datasets.
It is worth noting that even though LINE is designed for large-scale graphs, it takes a much longer time to converge (Figure \ref{fig:scale}). This is due to the fact that the number of iterations required by LINE for convergence is proportional to the number of edges~\cite{DBLP:line}. On the other hand, taking advantage of node attributes and uncertainty modelling, GLACE achieves convergence substantially faster. GLACE achieves a significant performance boost even after 1 minute of training.

\subsection{Visualization}

\begin{figure}[h]
    \begin{minipage}{\textwidth}
        \centering
        \includegraphics[width=0.8\linewidth]{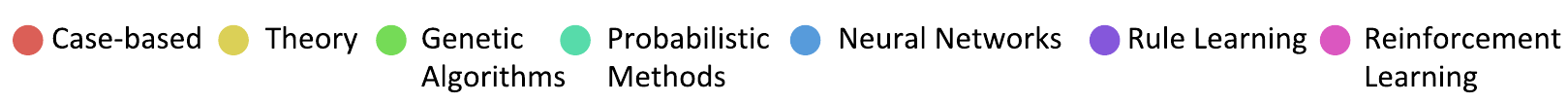}
    \end{minipage}
    \begin{minipage}{\textwidth}
        \centering
        \begin{subfigure}{0.24\textwidth}
            \includegraphics[width=\linewidth]{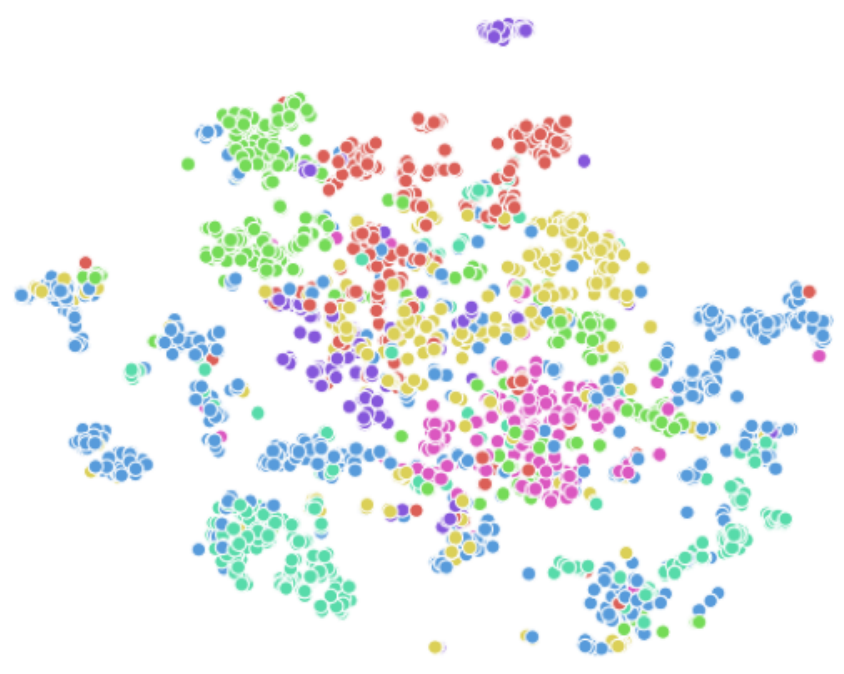}
            \subcaption{\small G2G}
            \label{fig:g2g}
        \end{subfigure}
        \begin{subfigure}{0.24\textwidth}
            \includegraphics[width=\linewidth]{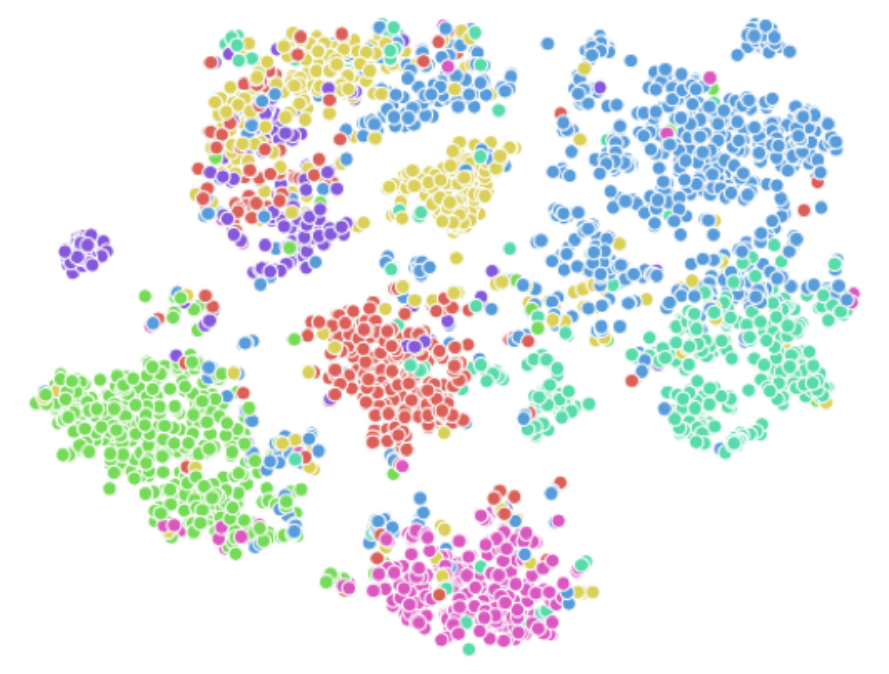} 
            \caption{\small LACE}
            \label{fig:glace_1}
        \end{subfigure}
        \begin{subfigure}{0.24\textwidth}
            \includegraphics[width=\linewidth]{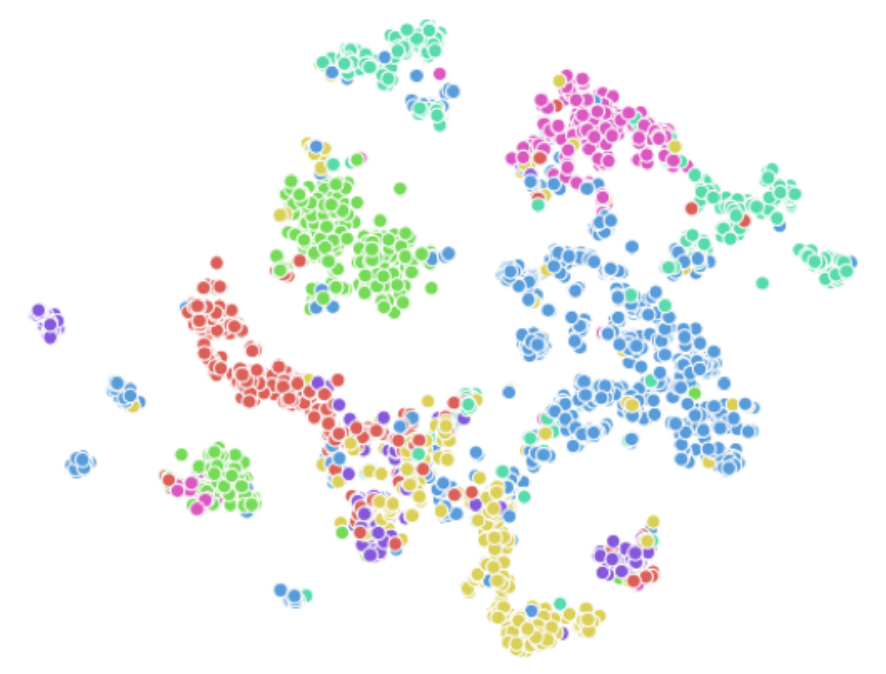} 
            \caption{\small GLACE}
            \label{fig:glace_2}
        \end{subfigure}
    \end{minipage}
    \caption{\small Visualization of Cora-ML graph (L = 64).}
    \label{fig:cora_ml_viz}
    \vspace{-5mm}
\end{figure}

We evaluate the ability to visualize the Cora-ML citation network. 
First, each model learns 128-dimensional node embeddings ($L=64$ for Gaussians). 
Then, the dimensions are reduced to 2 dimensions using t-SNE. 
Figure~\ref{fig:cora_ml_viz} shows the visualizations from the models which produced the best layouts. 
The color of a node (i.e.\ a paper) represents one of the seven research areas.
G2G produces moderately good clustering, but papers belonging to different areas are still not clearly separated. 
LACE and GLACE learn node embeddings that can clearly separate different classes. 
GLACE produces the best result in terms of tightly clustered papers of the same area with clearly visible boundaries.

\section{Conclusion}

We present GLACE, an unsupervised learning approach to efficiently learn node embeddings as probability distributions to capture uncertainty of the representations. GLACE learns from both node attributes and graph structural information, and is efficient, scalable and easily generalizable to different types of graphs. GLACE has been evaluated with respect to several state-of-the-art embedding methods in different graph analysis tasks, and the results demonstrate that our method significantly outperforms all the evaluated baselines. 

\bibliographystyle{splncs04}
\bibliography{bibliography.bib}

\end{document}